\documentclass[manuscript]{acmart}

\AtBeginDocument{%
  \providecommand\BibTeX{{%
    \normalfont B\kern-0.5em{\scshape i\kern-0.25em b}\kern-0.8em\TeX}}}

\copyrightyear{2021}
\acmYear{2021}
\setcopyright{acmlicensed}\acmConference[IUI '21]{26th International Conference on Intelligent User Interfaces}{April 14--17, 2021}{College Station, TX, USA}
\acmBooktitle{26th International Conference on Intelligent User Interfaces (IUI '21), April 14--17, 2021, College Station, TX, USA}
\acmPrice{15.00}
\acmDOI{10.1145/3397481.3450682}
\acmISBN{978-1-4503-8017-1/21/04}

\setcopyright{none}
\renewcommand\footnotetextcopyrightpermission[1]{}
\usepackage{wrapfig}
\usepackage[linesnumbered,ruled,]{algorithm2e}
\usepackage{subfig}
\usepackage{enumitem}

\newcommand{\ignore}[1]{}

\begin{document}

% For arxiv:

\title{Decision Rule Elicitation for Domain Adaptation}
\titlenote{{This is the pre-print version. The paper is published in the proceedings of IUI 2021 conference. Definitive version DOI: \url{https://doi.org/10.1145/3397481.3450682}.}}

\author{Alexander Nikitin}
\email{alexander.nikitin@aalto.fi}
%%\orcid{1234-5678-9012}
\affiliation{%
  \institution{Helsinki Institute for Information Technology HIIT, Department of Computer Science, Aalto University}
  %%\streetaddress{P.O. Box 1212}
  \city{Espoo}
  \country{Finland}
  %%\state{Ohio}
  %%\postcode{43017-6221}
}

\author{Samuel Kaski}
\email{samuel.kaski@aalto.fi}
\affiliation{%
  \institution{Helsinki Institute for Information Technology HIIT, Department of Computer Science, Aalto University}
  \city{Espoo}
  \country{Finland}
}
\affiliation{%
  \institution{Department of Computer Science, University of Manchester}
  \city{Manchester}
  \country{UK}
}

\newif\iffull
%\fullfalse
\fulltrue

\renewcommand{\shortauthors}{Nikitin and Kaski}

\begin{abstract}
Human-in-the-loop machine learning is widely used in artificial intelligence (AI) to elicit labels for data points from experts or to provide feedback on how close the predicted results are to the target. This simplifies away all the details of the decision-making process of the expert. In this work, we allow the experts to additionally produce decision rules describing their decision-making; the rules are expected to be imperfect but to give additional information. In particular, the rules can extend to new distributions, and hence enable significantly improving performance for cases where the training and testing distributions differ, such as in domain adaptation. We apply the proposed method to lifelong learning and domain adaptation problems and discuss applications in other branches of AI, such as knowledge acquisition problems in expert systems. In simulated and real-user studies, we show that decision rule elicitation improves domain adaptation of the algorithm and helps to propagate expert's knowledge to the AI model. 
\end{abstract}




\begin{teaserfigure}
\begin{center}
  \includegraphics[width=\textwidth]{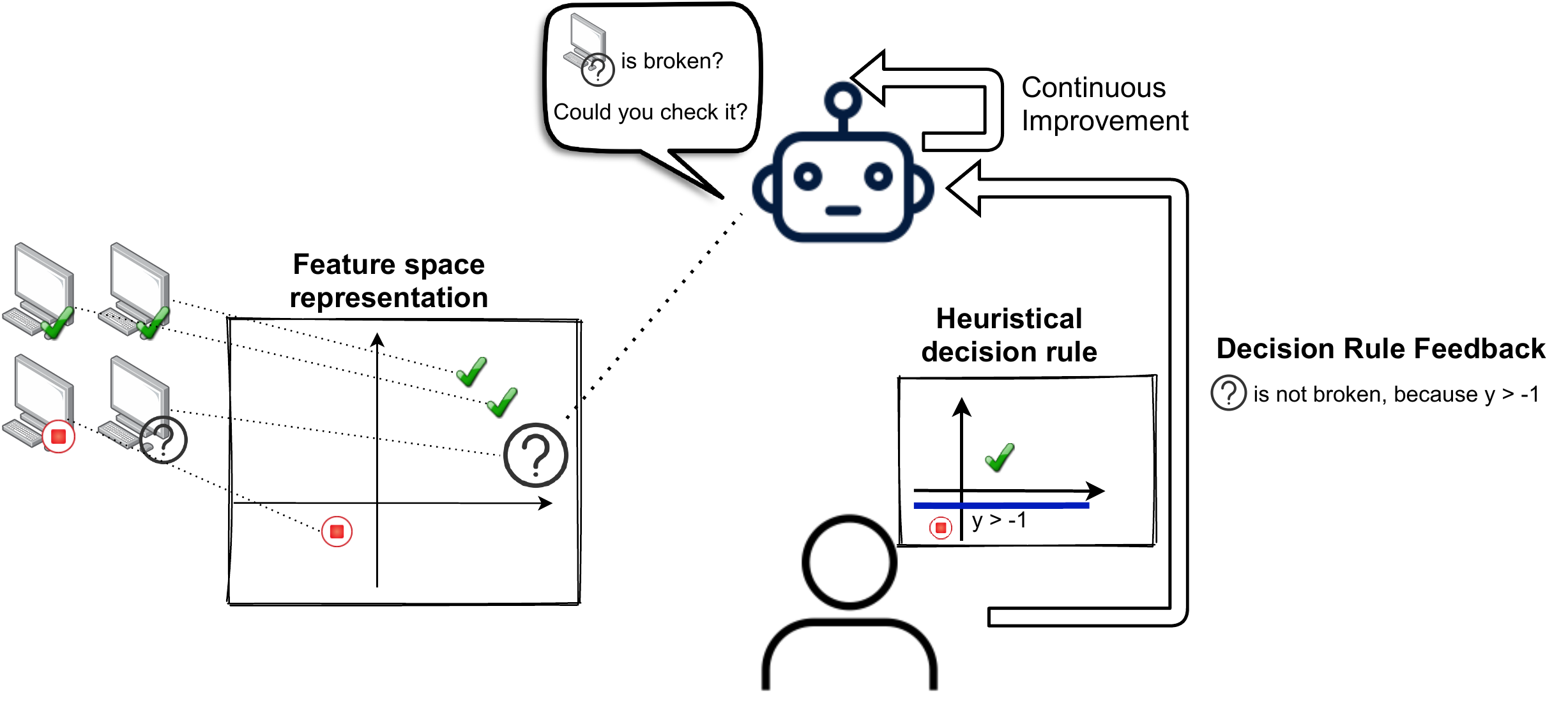}
  \caption{We consider a task where a machine learning model has been trained to predict breaks in a set of workstations. When the system receives new data about the workstations, it predicts their fault-risk and passes some of the predictions to the expert user for evaluation. Instead of just telling whether they agree, as in current human-in-the-loop systems, the expert gives a heuristic rule describing their decision-making. The machine learning system uses this rule to improve its decision making. Even though the rules will be imperfect and noisy, they will incorporate additional information available to the human users but not the system.}
  \label{fig:teaser}
\end{center}
\end{teaserfigure}
\maketitle
\section{Introduction}
\label{section:introduction}
This work is inspired by practical needs in industry, in equipping maintenance systems with machine learning assistance. Straightforward "black-box" machine learning approaches would produce a prediction, for instance, a binary variable indicating there may be something wrong with the workstations in 
Figure \ref{fig:teaser}. Human-in-the-loop approaches accept feedback from human experts in the form of corrected values. Experience with real experts in practical maintenance systems quickly reveals that the experts have their own rules of thumb, by which they make their maintenance decisions. While working with a telecom company, we noticed, to our surprise, that not only were the experts able to explicate the rules, the rules also improved the performance of the model and showed good generalization performance.

Knowledge from domain experts has been included in automated systems at least since the expert systems popular in the '80s. Difficulty in eliciting sets of consistent rules from experts was one of the most important difficulties \cite{hart1985knowledge} explaining why expert systems are not popular anymore. Another widely-used formulation for human input is prior elicitation, that is, getting the prior distributions of model parameters from experts for Bayesian inference \cite{garthwaite2013prior,kadane1980interactive}. The elicitation can also be done indirectly, by eliciting similarities, for instance, \cite{afrabandpey2017interactive}. 

Even though humans may be able to explain their decisions using formal decision rules, using logical predicates, that is laborious and error-prone. Psychological research of decision-making has shown that fast heuristics are often useful for generalization, one of the most popular branches being fast and frugal decision-making \cite{gigerenzer1999simple}. Essentially, fast and frugal trees are a shallow decision tree that provides a heuristic for decision-making. Experimental studies have shown that fast and frugal heuristics can be used as a fast alternative to fully-Bayesian decisions. In this work, we build on the assumption that human experts either naturally use such trees, or at least can easily generate them, and use them as inputs to machine learning.

In this paper, we introduce a new way to elicit and use the experts' heuristic rules to improve machine learning. We assume the explicit feedback can be expressed as a Boolean formula, and handle it as an output of a weak learner \cite{schapire1990strength}. The rules are elicited for wrongly predicted data samples.
Additionally, we show that this approach helps to deal with several modern artificial intelligence problems where generalization to new distributions is required: out-of-distribution generalization, lifelong learning, and learning from the feedback. 

\section{Related work}
\label{section:background}
This work extends directly earlier work in human-in-the-loop machine learning or human-centered machine learning \cite{Amershi_Cakmak_Knox_Kulesza_2014}. The new type of human input we introduce could be directly used in active learning  \cite{lewis1994} instead of the usual labels requested from users to unlabeled samples.

In reinforcement learning \cite{DBLP:conf/aaai/MandelLBP17}, users can give feedback to a system on the quality of its performance. In \cite{christiano2017deep}, an agent's goals were derived from the preferences of non-expert humans on trajectories. Cooperative inverse reinforcement learning (CIRL) \cite{hadfield2016cooperative} is a related approach where a robot and a human user collaborate in a partial-information game to maximize the human’s reward function.

An expert system is an artificial intelligence system that emulates the decision-making process of an expert \cite{russell2002artificial}. The decision-making is expressed as procedural code or simple decision rules (logical predicates). Building of the necessary knowledge base for an expert system is called knowledge engineering \cite{russell2002artificial}. Knowledge acquisition is one of the significant and critical tasks in developing an expert system \cite{muhammad2018problems}. \ignore{The knowledge acquisition process consists of the following stages: knowledge elicitation, interpretation, and structuring \cite{studer1998knowledge}.} Our work could help to construct the databases, but has not been designed to solve the critical problem of expert systems, that the rules may be inconsistent.

A recent line of works, complementary to ours, has done interactive knowledge elicitation for statistical prediction tasks. In \cite{daee2017knowledge}, the authors proposed a method of knowledge elicitation for high-dimensional datasets, where an expert knows about the relevance of the covariates or values of the regression coefficients. A notable example of the practical knowledge elicitation applications for genomics prediction was proposed in \cite{sundin2018improving}.
 
We elicit simple heuristics from the human experts, and the rules must be natural and easy for the experts to express. The research line on fast and frugal trees gives insights to this \cite{goldstein2002models}. 

\section{The proposed method}
\label{section:theory}
\subsection{Problem definition}
\label{section:formal_definition}
We consider a prediction problem where an output $y^{i}_{train} \in Y$ needs to be predicted based on an input $X^{i}_{train} \in \mathcal{X}$. The prediction model $\mathcal{M}: \mathcal{X} \rightarrow Y$ is trained using a historical dataset $(X_{train}, y_{train})$. The testing data comes as a sequence of sets $X^{t}_{test}, \forall t \in [1, T]$; the data can be out-of-distribution or from another domain. We consider the setting where the model $M$ is trained once and should serve for a long time, improving the results "on the fly." This paradigm is often called lifelong machine learning \cite{chen2018lifelong}. The user $U$ continuously interacts with the model in the following way, while the system is in operation:
\begin{itemize}[topsep=0pt]
    \item test data $X_{test}^{t}$ arrive at time $t$, and the model produces a prediction for each item,  $\hat{y}^t_{test} = \mathcal{M}(X^t_{test})$,
    \item $U$ evaluates the predictions for a subset of $X^t_{test}$ chosen by the user of the model, and called here the \emph{observable dataset}. U finds the mistakes made by the model $\mathcal{M}$, and provides feedback $f$ on them. We will not consider further in this paper how the observable data set is chosen; a natural choice would be based on the upper confidence bound, for instance,
    \item $\mathcal{M}$ uses the feedback $f$ for continuous learning and self-improvement.
\end{itemize}

This process is described in more detail with Pseudocode~{\iffull \ref{listing:lifelong_with_expert} \else 1\fi} in Appendix~{\iffull \ref{appendix:training_the_algorithm} \else A\fi}.
In the following sections, we will consider only one testing round, but the method naturally generalizes to multiple testing rounds.

One of the key observations in the psychological literature is that the heuristics that experts would use to describe their decision-making heuristics can be described as fast and frugal trees and consequently by a logical predicate \cite{martignon2003naive}.  In the following sections, we show how to model this type of feedback and how it can be applied to natural language processing.

\subsection{User-aware prediction algorithm}
\label{section:algorithm}
We consider each of the decision rules, extracted from the users' feedback, as a weak learner \cite{schapire1990strength}. The combination of the learners often performs better than each weak learner; this is the main reason for the popularity of the algorithms such as gradient boosting and random forests. We combine the $f_i$ to
$$C_{feedback}(x) = \sigma(\sum_{i=1}^{F} f_i(x) sim(X_{test}^{i}, x) \theta_i)$$
where $f_i$ the decision rule extracted from the $i$th feedback, $F$ is the total number of the decision rules, $sim$ is a measure of similarity between test example (where the feedback is given) and $x$, $\theta_{i}$ is the weight of the $i$th classifier learned from the data, $\sigma$ is an activation function, $\mathcal{C}_{feedback}$ and the $\mathcal{C}_{hist}$ (a classifier that is trained on $(X_{train}, y_{train})$) are parameterized by $\theta_{f}$ and $\theta_{hist}$.

We combine the human feedback-derived and data-derived classification algorithms with a weighted combination: 
$$\mathcal{C}(x) = \alpha \mathcal{C}_{hist}(x, \theta_{hist}) + (1 - \alpha) \mathcal{C}_{feedback}(x, \theta_{f})$$
Here $\alpha$ is a weight that is used for making the algorithm more or less sensitive to the human feedback. Alternatively, it could be learned in the same manner as the parameters $\theta_i$ in the following subsection. 

This algorithm could be extended to
$$\mathcal{C}(x) = \alpha(x) * \mathcal{C}_{hist}(x, \theta_{hist}) + (1 - \alpha(x)) * \mathcal{C}_{feedback}(x, \theta_{f})$$
where $\alpha$ is variable across different data samples. For simplicity, we will not study selection of $\alpha$ in this work, and will consider $\alpha$ as a hyperparameter of the algorithm and constant for all the samples.

\subsubsection{Training the algorithm}
\label{section:training_the_algorithm}
Having the expressions for the gradients of the loss, one can apply one of the standard optimization methods to estimate the parameters $\theta$, e.g., stochastic gradient descent (SGD) \cite{kiefer1952stochastic}. We derive the gradients for binary classification problem with binary cross-entropy loss function in Appendix~{\iffull \ref{appendix:training_binary_classification} \else A.1\fi}.

We can consider two cases: where the initial model can be changed and where it should remain. In the first case, while algorithm training, the parameters $\theta_{hist}$ and $\theta_f$ will be updated via an optimization algorithm. And, in the second case, an optimization algorithm will update only the parameters $\theta_f$. 

\subsection{Modeling the experts for the experiments}
\label{section:modeling_the_humans}
In this section, we propose a method for human modeling that can be used for studies when experimenting with real domain experts is too difficult for various reasons (cost, time, etc.). The modeled humans have several properties to reflect the reality adequately. The \textit{depth} of the human is the tendency of human to construct complex rules. The second parameter is \textit{experience} and it can be defined in two following ways:
\begin{enumerate}[topsep=0pt]
    \item fraction of the observed data (from the mixture of all distributions),
    \item if we consider the data as a mixture of a number of distributions $p_1, ..., p_n$ and the human has faced the data from the distributions $p_{i_1}, ..., p_{i_k}$, then we will call value $\frac{k}{n}$ as human experience.
\end{enumerate}
Thus, human feedback decision rules can be modeled as a set of decision trees \cite{breiman1984classification} that trained on the historical data (observed from all or a part of the distributions). The decision trees fitted to the data using CART algorithm \cite{breiman1984classification}. Parameter depth shows the complexity of the rules; our experiments show that there is no need for too complex rules. Experience refers to the variability of the samples seen by the expert.

\section{Experiments}
\label{section:experiments}

In this section, we analyze the experimental performance of the proposed methods. We first demonstrate the method with simulated humans, modelled to produce decision trees (section \ref{section:experiment_synthetic_switch_user_modelling}), and then carry out several user studies (Sections \ref{section:experiment_user_study_switching} and \ref{section:experiment_sentiment_analysis}).

\subsection{Switching distribution and user modelling}
\label{section:experiment_synthetic_switch_user_modelling}

\begin{wrapfigure}{r}{4cm}
\begin{center}
    \includegraphics[width=4cm]{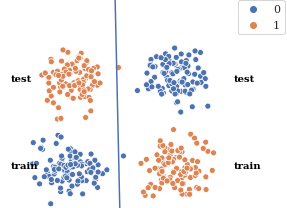}
  \end{center}
\caption{Switching distribution}
\label{figure:experiment_distr_switches}
\end{wrapfigure}

In this experiment, we consider the dataset $X_{train}, y_{train}, X_{test}, y_{test}$ presented in Figure \ref{figure:experiment_distr_switches}. The total size $N=400$, and each cluster contains 100 points.
Details are in Appendix~{\iffull \ref{appendix:switching_distribution} \else B.1\fi}.

\par We simulate 10 human experts as follows. Human depth $d\sim U([1, 2, 3, 4])$, human experience $e \sim N(0.3, 0.05)$ indicating the proportion of the data seen by the human. We train the decision rule for each human as the decision trees of depth $d$ on $e$ part of the data generated from the mixture of train and test distributions. Thus, we model the experts as experienced across all the distributions, however, having simple heuristic rules. The decision trees were fitted to the data using CART algorithm \cite{breiman1984classification}.

We collected feedback decision rules from 40 modeled experts. Then, we compared the data-driven model, the model with decision rule feedback, and the model with new labeled samples as feedback. 
\begin{table*}[h!]
    \caption{Experimental evaluation on synthetic dataset with user modeling (Accuracy)}
    \label{table:user_study_synthetic_main}
    \centering
    \begin{tabular}{ | c | c | c | c | }
         \hline
         feedback & Train distr. & Test distr. & Train + test distr.\\
        \hline 
         Decision rule feedback & \textbf{1.0} & \textbf{0.88} & \textbf{0.94} \\
        \hline
         Labels feedback 40 experts & 0.0 & 1.0 & 0.5 \\ 
        \hline
        No feedback & 0.995 & 0.005 & 0.5\\
        \hline
    \end{tabular}
\end{table*}

We used similarity function $sim \equiv constant = 1$ and sigmoid activation function $\sigma(x) = \frac{1}{1 + e^{-x}}$.
\par What is striking in this table, is the high accuracy for the method with decision rule feedback for train distribution and test distribution. That confirms the domain adaptation properties of the proposed method for modeled decision rules.

\subsection{User studies}
\subsubsection{Switching distribution}
\label{section:experiment_user_study_switching}
We conducted a user study on a dataset similar to the one studied in Section~\ref{section:experiment_synthetic_switch_user_modelling} using the same data generation process. We followed the proposed algorithm of collecting the users' feedback responses. We asked 10 people to formulate the decision rules on how to distinguish the two classes. After this step, we ran the logistic regression model on the dataset and evaluated it on the small subsets of the test data. Misclassified data points were propagated to the users, who could both use the previously formulated rule to provide the feedback, or if the rule did not classify a new point correctly, they were able to modify the rule.
\par We fix $\alpha(x) = 0.5$ and found $\theta$ using SGD as described in Section \ref{section:training_the_algorithm}.
\par The decision rule feedback model with the rules from all 10 participants improved the test data results compared to data-driven logistic regression from 0\% to 99.8\%, without significant performance decrease on training distribution.
The majority of the rules had depth 2. Experts with low experience tended to create overcomplicated rules and often changed their rules while acquiring new data points; at the same time, experts with high experience created simple and precise rules.

\subsubsection{Sentiment analysis}
\label{section:experiment_sentiment_analysis}
In this user study, we evaluated domain adaptation capability of the proposed framework on a sentiment analysis task. Essentially, it is a binary text classification task with two classes: positive and negative. The goal is to understand the polarity of the given text. We trained a logistic regression model on the movie review dataset \cite{movie_review_dataset}. We evaluated the model on topics from a new domain, different from movies. For this purpose, we used multiple domain sentiment analysis dataset \cite{domain_adaptation_for_sentiment}. The topics in the second dataset are much broader than in the movie review dataset, including music, video, electronics, grocery, software. The test dataset contained 25 topics, and 10 people participated in the experiment.

We tested the proposed framework with the following sequence of actions. Each participant of the user study received the reviews one by one. For each review, the right class is known, and the predicted class is also known. The participants needed to formulate a rule that helps to predict the correct class of the review. The participants often used simple predicates such as the presence of a particular phrase in the text, but sometimes also more sophisticated rules such as regular expressions. In total, the participants generated 139 rules for the positive class and 142 for the negative.

\begin{figure}
\centering
\subfloat[]{
  \includegraphics[width=60mm]{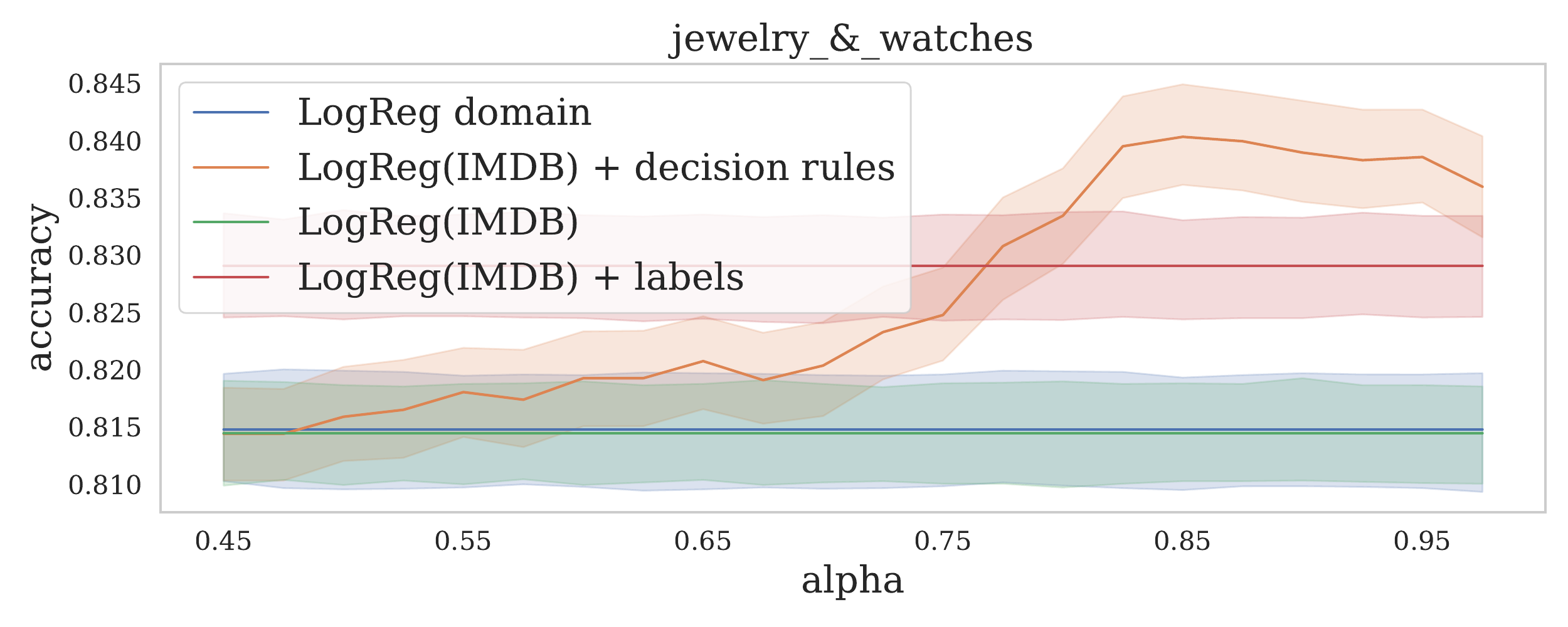}
}
\hspace{0mm}
\subfloat[]{
  \includegraphics[width=60mm]{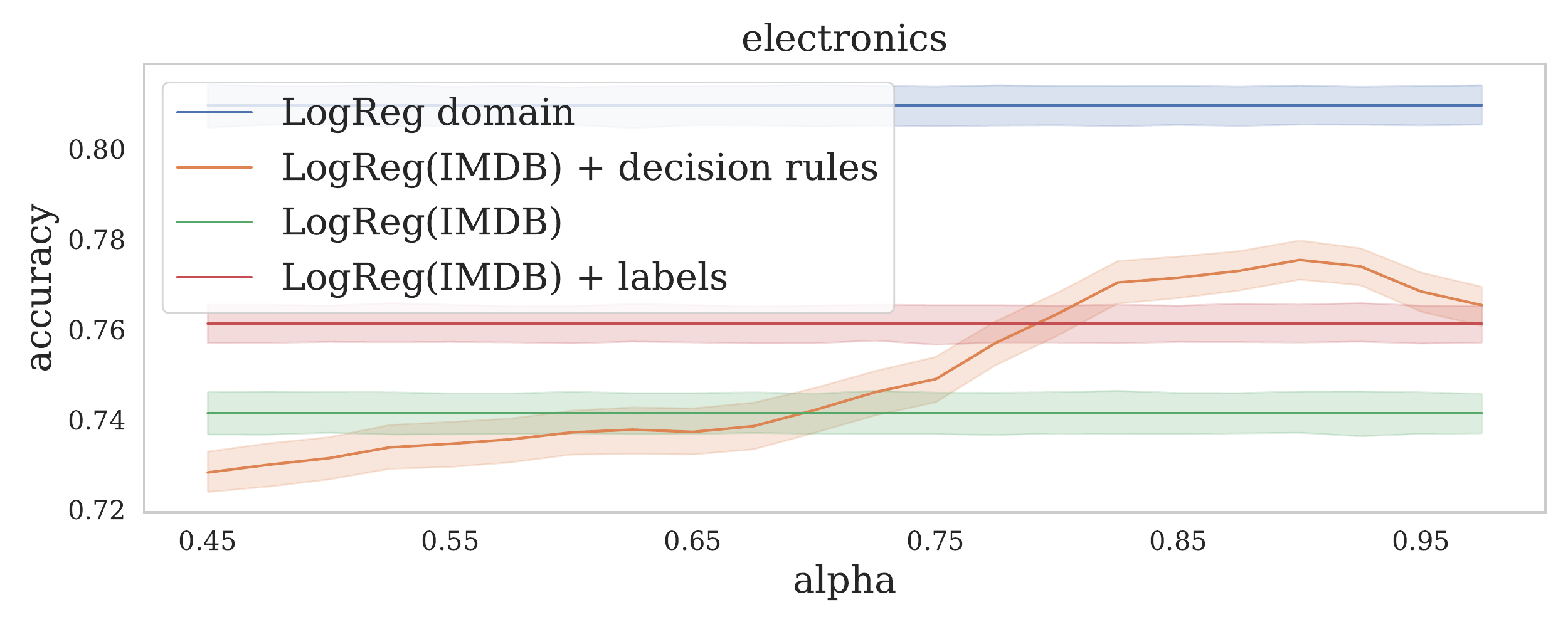}
}
\hspace{0mm}

\caption{Model with decision rule feedback performs better than label feedback (topics: "jewelry and watches" and "electronics"). \emph{LogReg(IMDB)} is logistic regression trained on IMDB; \emph{LogReg domain} is logistic regression trained on the considered domain, \emph{(LogReg(IMDB) + labels)} is logistic regression trained on IMDB with feedback as labels, \emph{(LogReg(IMDB) + decision rules)} is logistic regression with decision rule feedback (our proposed method).}
\label{figure:experiment_sentiment_analysis_short}
\end{figure}

Experimental evaluation showed that the accuracy of the model with decision rule feedback was better for most domains than the accuracy of the model without feedback or with labels as feedback. Two selected topics are presented in Figure \ref{figure:experiment_sentiment_analysis_short}. We can observe that our method outperforms the data-driven model and the model with labels as feedback for $\alpha \approx 0.85$. We also can see that for the topic "jewelry\_\&\_watches" the proposed approach performs even better than a model specifically trained on this domain. The complete set of the results for this experiment, as well as the experimental setting, are presented in Appendix~{\iffull \ref{appendix:sent_analysis_and_user_studies} \else B.4\fi}.

\section{Conclusion and discussion}
\label{section:conclusion}
In this work, we proposed a method for eliciting decision rules from experts, and for using them effectively for machine learning. The proposed method was demonstrated by simulations and user experiments to be useful for domain adaptation. It is also less laborious than the standard knowledge elicitation process. An additional advantage of the method is that it is model-agnostic and can be used for lifelong learning. The experiments also showed that the method helps an algorithm to generalize better to a new distribution, in a setting with a dramatic change in the distributions. In a sentiment analysis experiment, we showed the proposed method's applicability in a real-life problem, and demonstrated its domain generalization capability across different topics.

\par An important restriction of this work is that the method is applicable only when experts are able to elicit heuristic decision rules. However, we believe that decision rule elicitation is possible for the majority of the modern ML problems. 
\par Decision rule elicitation gives a second wind to the expert systems by resolving a challenging knowledge acquisition problem; it may lead to more frequent usage of human-crafted decision rules in practice. This required a statistical treatment of the rules, acknowledging that they are imperfect. The proposed method is also an essential step in human-in-the-loop methods; we show that humans can be more effectively included in the loop.

\par We believe that the proposed method is an essential step towards more efficient interaction between experts and a model. As a generalization of our idea, one can consider providing the feedback to the model in natural language, where instead of using the decision rule predicates, an expert provides an explanation of their decision in raw text. This idea needs more careful investigation, and our method is a good indication that it is possible in principle.

\begin{acks}
This work was supported by the Academy of Finland (Flagship program: Finnish Center for Artificial Intelligence FCAI) grants 319264 and 292334. We acknowledge the computational resources provided by the Aalto Science-IT Project.
\end{acks}

\bibliographystyle{ACM-Reference-Format}
\bibliography{main}

\iffull

\appendix
\section{Training the algorithm}
\label{appendix:training_the_algorithm}
In the listing \ref{listing:lifelong_with_expert}, we present a pseudocode describing the training loop in the setup we used.

\begin{algorithm}[H]
\SetAlgoLined
 M = train\_ml\_model(historical\_data)\;
 \While{The system is working}{
  test\_data = gather\_latest\_test\_data()\;
  predictions = M.predict(test\_data)\;
  \For {user in all\_users}{
   feedback\_rules = user.use\_predictions(predictions)\;
   M.self\_improve(feedback\_rules)\;
   }
 }
 \caption{Lifelong machine learning problem with the experts' feedback}
 \label{listing:lifelong_with_expert}
\end{algorithm}

\subsection{Binary classification}
\label{appendix:training_binary_classification}
Let's consider a binary classification task with cross-entropy as a loss function:
$$\mathcal{L} = -y_i log(C(x_i)) - (1 - y_i) log(1 - \mathcal{C}(x_i))$$

The gradients for the parameters $\theta_{hist}$ and $\theta_{f}$ could be written in the following form:
$$\frac{\partial L}{\partial \theta_{hist}} = \alpha \frac{\partial C_h}{\partial \theta_h} (\frac{1 - y_i}{1 - C(x_i)} - \frac{y_i}{C(x)})$$
$$\frac{\partial L}{\partial \theta_{f}} = (1 - \alpha) \frac{\partial C_f}{\partial \theta_f} (\frac{1 - y_i}{1 - C(x_i)} - \frac{y_i}{C(x)})$$

\subsection{Estimation of the mixture coefficient $\alpha$}
\subsubsection{Mixture coefficient between the model and the feedback}
An essential practical aspect of the algorithm is the right choice of the averaging coefficient between the machine learning model and the model based on users' decision rules. We propose various strategies for choosing $\alpha$:
\begin{enumerate}
    \item setting $\alpha$ as a hyperparameter of the algorithm,
    \item choosing larger coefficient for human-based decision rules in the areas of $\mathcal{X}$ of the higher uncertainty,
    \item non-constant $\alpha$ dependent on probability density, can be estimated using kernel density estimation (KDE),
    \item compare training and testing distribution with two-sample testing.
\end{enumerate}
Experimentally we will focus only on setting $\alpha$ as a hyperparameter. We leave hyperparameter choosing for future research.

\section{Experiments}

\subsection{Switching distribution}
\label{appendix:switching_distribution}
Here we provide extensive details on data generation process for the experiments described in Sections \ref{section:experiment_synthetic_switch_user_modelling} and \ref{section:experiment_user_study_switching}.

The training dataset is a set of points described by two features of total size $N=200$, with $\frac{N}{2}$ points of class $1$ ($y_i=0$) and $\frac{N}{2}$ points of class $0$ ($y_i=0$) by the following rules
\[
\left\{
\begin{array}{ll}
x^{train}_i \sim N(\mu_1, \Sigma_1) & \mbox{if $y_i=1$},\\
x^{train}_i \sim N(\mu_2, \Sigma_2) & \mbox{if $y_i=0$}.\\
\end{array}
\right.
\]
The testing dataset $X_{test}, y_{test}$ is given from other distribution in order to emulate how the human decision rules feedback help to better generalize model to out-of-distribution samples. $X_{test}, y_{test}$ of size $N=200$, with $\frac{N}{2}$ points of class $1$ ($y_i=0$) and $\frac{N}{2}$ points of class $0$ ($y_i=0$) by the following rules
\[
\left\{
\begin{array}{ll}
x^{test}_i \sim N(\mu^\prime_1, \Sigma^\prime_1) & \mbox{if $y_i=1$},\\
x^{test}_i \sim N(\mu^\prime_2, \Sigma^\prime_2) & \mbox{if $y_i=0$}.\\
\end{array}
\right.
\]

For the experiments we took $\mu_1 = \begin{pmatrix} 0 & 0 \end{pmatrix}$, $\mu_2 = \begin{pmatrix} 6 & 0 \end{pmatrix}$, $\mu_1^\prime = \begin{pmatrix} 6 & 6 \end{pmatrix}$, $\mu_2^\prime = \begin{pmatrix} 0 & 6 \end{pmatrix}$,
$\Sigma_1 = \Sigma_2 = \Sigma_1^\prime = \Sigma_2^\prime = \begin{pmatrix} 1 & 0 \\ 0 & 1 \end{pmatrix}$.
\subsection{Synthetic dataset and user modeling}
\label{appendix:synth_dataset_and_user_modeling}
In this section, we present full results for the experiments with switching dataset and modeled experts. We generated humans as described in the Section \ref{section:modeling_the_humans}. We used logistic regression as a base model, found misclassified samples, and provided feedback decision rules from all experts. The number of generated human experts is provided in the column n\_humans. Each modeled expert can produce the feedback in two following ways: (1) if its decision rule can classify a particular sample, then it returns it's decision rule, (2) if this sample cannot be classified by its decision rule, it adds new sample to its experience, train the underlying decision tree again, and returns it as a feedback decision rule.
\begin{table}[h!]
    \caption{User modeling results on synthetic dataset}
    \label{table:user_study_synthetic}
    \centering
    \begin{tabular}{ | c | c | c | c |  c | c | c | c | }
         \hline
         n\_humans & n\_feedback & \multicolumn{3}{c}{LogReg + decision rule feedback} & \multicolumn{3}{c|}{LogReg + label feedback}\\
         \hline
          &  & Train distr & Test distr & Train + test distr & Train distr & Test distr & Train + test distr\\ 
         \hline
         1 & 8 & 0.705 & 0.5 & 0.603 & 0.965 & 0.495 & 0.73 \\  
         \hline
         5 & 33 & 0.96 & 0.47 & 0.715 & 0.975 & 0.495 & 0.735 \\  
         \hline
         10 & 67 & 0.99 & 0.65 & 0.82 & 0.28 & 0.5 & 0.593 \\  
         \hline
         20 & 133 & 1.0 & 0.45 & 0.725 & 0.015 & 0.78 & 0.398 \\  
         \hline
         30 & 241 & 0.995 & 0.565 & 0.78 & 0.0 & 1.0 & 0.5 \\  
         \hline
         40 & 288 & \textbf{1.0} & \textbf{0.88} & \textbf{0.94} & 0.0 & 1.0 & 0.5 \\ 
         \hline
         50 & 385 & 1.0 & 0.81 & 0.905 & 0.0 & 1.0 & 0.5 \\ 
         \hline
    \end{tabular}
\end{table}

\subsection{Synthetic dataset and user studies}
\label{appendix:synth_dataset_and_user_studies}
We describe the profiles of the experts in Table~\ref{table:user_study_synthetic_profiles}.

\begin{table}[H]
    \caption{The profiles of the experts participating in the user study with synthetic dataset}
    \label{table:user_study_synthetic_profiles}
    \centering
    \begin{tabular}{ | c | c | c | c | }
         \hline
         User & Experience & n\_feedback\\ 
         \hline
         0 & 0.1 & 10 \\  
         \hline
         1 & 0.1 & 10 \\
         \hline
         2 & 0.1 & 9 \\ 
         \hline
         3 & 0.03 & 8 \\ 
         \hline
         4 & 0.01 & 9 \\ 
         \hline
         5 & 0.01 & 9 \\ 
         \hline
         6 & 0.03 & 9 \\ 
         \hline
         7 & 0.03 & 9 \\ 
         \hline
         8 & 0.01 & 10 \\ 
         \hline
         8 & 0.03 & 10 \\ 
         \hline
    \end{tabular}
\end{table}

We also provide the user studies results for all the experts and evaluate our method with feedback decision rules elicited from each expert. Full results of the user study are presented in Table~\ref{table:user_study_synthetic}.

\label{appendix:table_synth_dataset_and_user_studies}
\begin{table}[H]
    \caption{Full results of the user study on the synthetic data}
    \label{table:full_user_study_synthetic}
    \centering
    \begin{tabular}{ | c | c | c | c | c | c | }
         \hline
         Model & User & $\alpha$ & Train distr. accuracy & Test distr. accuracy & Train + Test distr. accuracy\\ 
         \hline
         LogReg & No & - & 1.0 & 0.0 & 0.5 \\  
         \hline
         LogReg + decision rule feedback & 0 & 0.3 & 0.968 & 0.99 & 0.98 \\ 
         \hline
         LogReg + decision rule feedback & 1 & 0.3 & 0.995 & 1.0 & 0.998 \\  
         \hline
         LogReg + decision rule feedback & 2 & 0.3 & 0.980 & 1.0 & 0.988 \\ 
         \hline
         LogReg + decision rule feedback & 3 & 0.3 & 0.996 & 0.995 & 0.995 \\
         \hline
         LogReg + decision rule feedback & 4 & 0.3 & 0.995 & 0.015 & 0.505 \\
         \hline
         LogReg + decision rule feedback & 5 & 0.3 & 1.0 & 0.47 & 0.735 \\
         \hline
         LogReg + decision rule feedback & 6 & 0.3 & 0.99 & 0.985 & 0.988 \\ 
         \hline
         LogReg + decision rule feedback & 7 & 0.3 & 0.985 & 1.0 & 0.993 \\ 
         \hline
         LogReg + decision rule feedback & 8 & 0.3 & 0.995 & 0.995 & 0.995 \\  
         \hline
         LogReg + decision rule feedback & 9 & 0.3 & 0.98 & 0.5 & 0.743 \\  
         \hline
        \textbf{LogReg + decision rule feedback} & All & 0.3 & 0.995 & 0.998 & 0.743 \\  
         \hline
        LogReg + label feedback & All & - & 0.005 & 0.995 & 0.5025 \\  
         \hline
    \end{tabular}
\end{table}

\subsection{Sentiment analysis and user studies}
\label{appendix:sent_analysis_and_user_studies}

We conducted a user study with 10 human experts on the sentiment analysis problem. The algorithm was trained on IMDB dataset, and then evaluated on the different topics.
\par In Figures \ref{figure:sentiment_experiments_1}, \ref{figure:sentiment_experiments_2}, \ref{figure:sentiment_experiments_3}, and \ref{figure:sentiment_experiments_4} we compare accuracy (fraction of correctly classified samples among all samples) for 4 different methods:
\begin{enumerate}
    \item Logistic regression trained on IMDB (LogReg (IMDB)),
    \item Logistic regression train on a given domain (LogReg domain),
    \item Logistic regression with feedback as labels (LogReg(IMDB) + labels),
    \item The proposed method: (LogReg (IMDB) + decision rules).
\end{enumerate}
In Figures \ref{figure:sentiment_experiments_1}, \ref{figure:sentiment_experiments_2}, \ref{figure:sentiment_experiments_3}, and \ref{figure:sentiment_experiments_4}  we provide the results of the study for 25 different topics. The filled area is the standard error, and we estimated it by 40 bootstrap iterations. We can see that in 4 cases, our method outperforms all the other approaches for a certain value of $\alpha$. It also performs not worse than feedback as labels in the majority of the cases.
\par The generated rules vary from the simplest checks that a particular pattern occurs in the sentence (e.g., a predicate \texttt{"funny" in text}) to quite sophisticated rules involving regular expressions (e.g., a regular expression \texttt{* it('s | is ) the best .*}).
\par From this experiment, we also see that the correct choice of a hyperparameter $\alpha$ is vital for the empirical performance of the algorithm. We also can see that optimal values of $\alpha$ are close across the domains, which simplifies the usage of the algorithm. 

\begin{figure}[H]
\centering
\subfloat[]{
  \includegraphics[width=65mm]{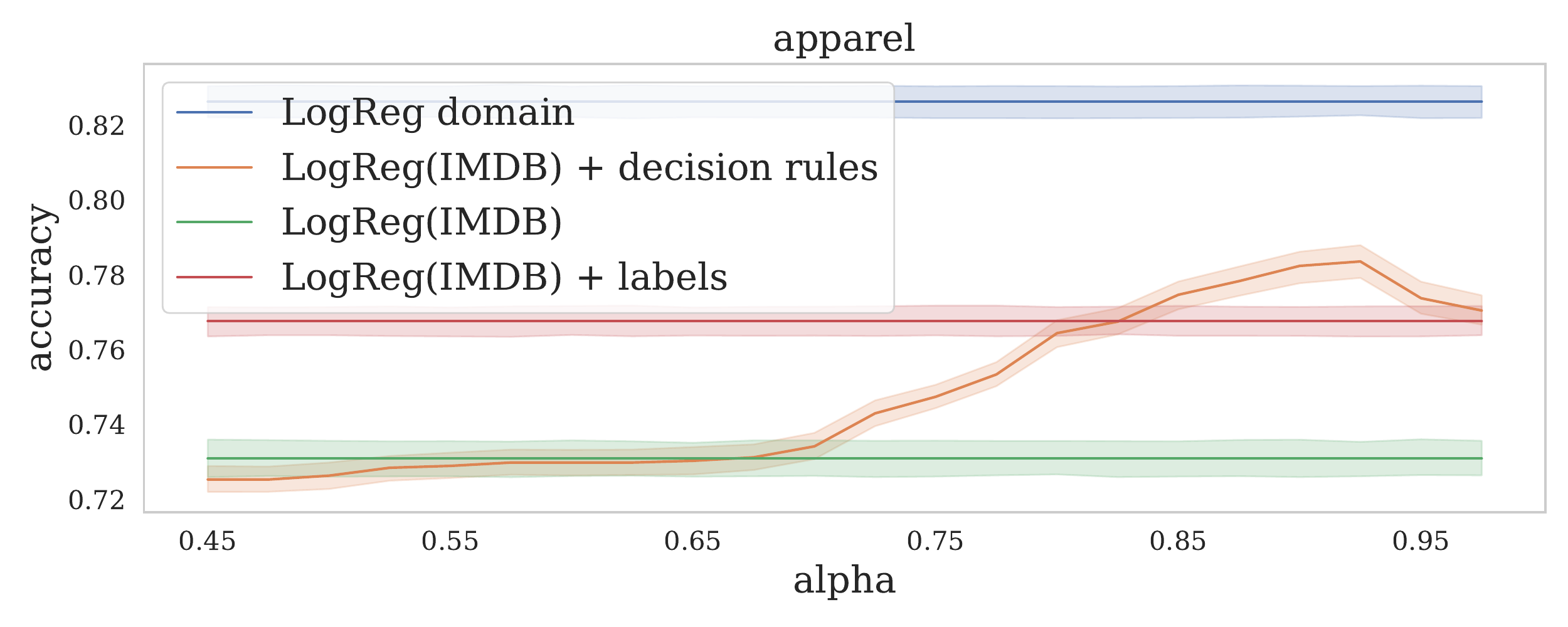}
}
\subfloat[]{
  \includegraphics[width=65mm]{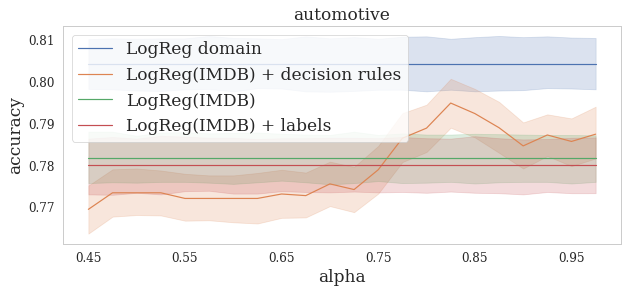}
}
\hspace{0mm}
\subfloat[]{
  \includegraphics[width=65mm]{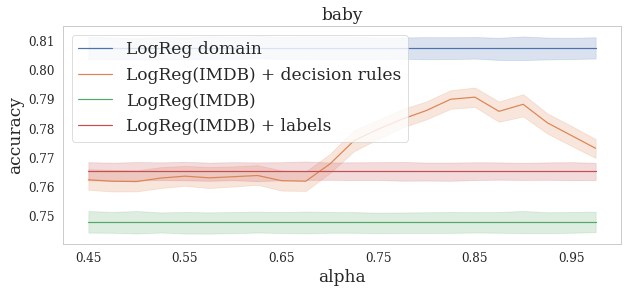}
}
\subfloat[]{
  \includegraphics[width=65mm]{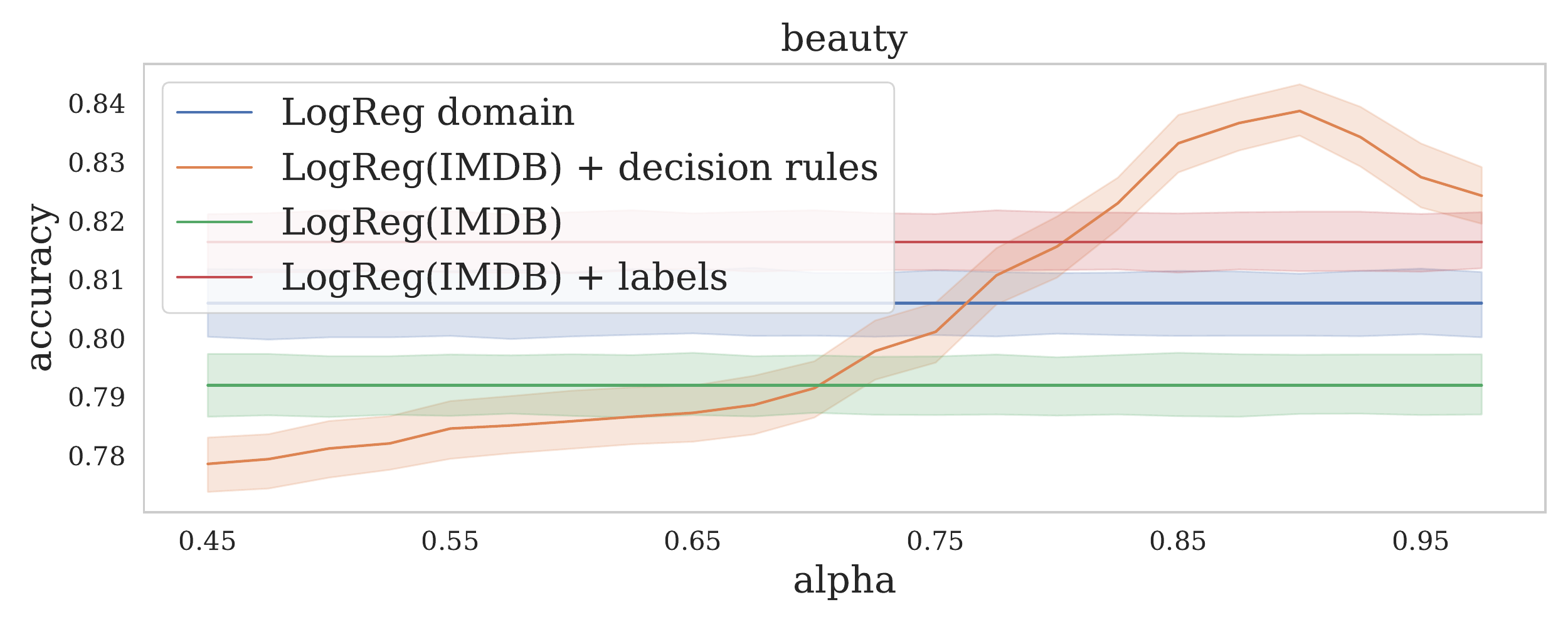}
}
\hspace{0mm}
\subfloat[]{
  \includegraphics[width=65mm]{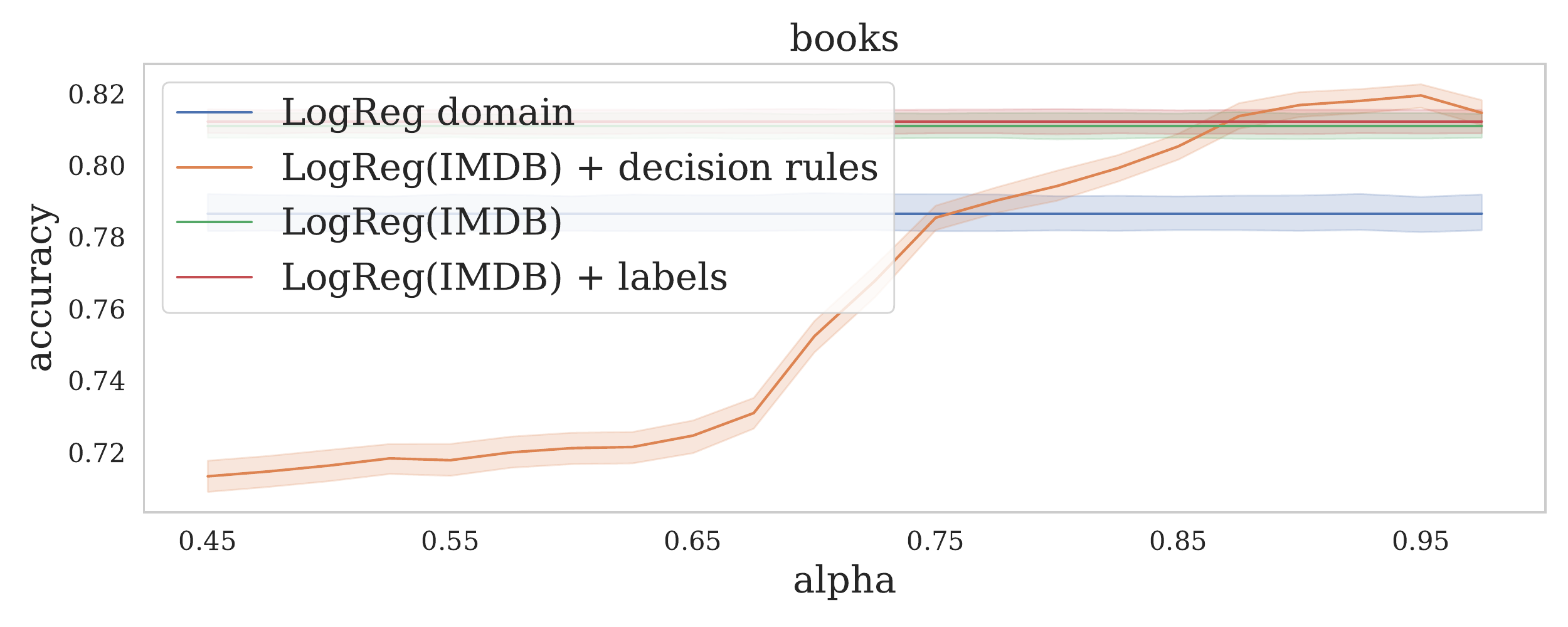}
}
\subfloat[]{
  \includegraphics[width=65mm]{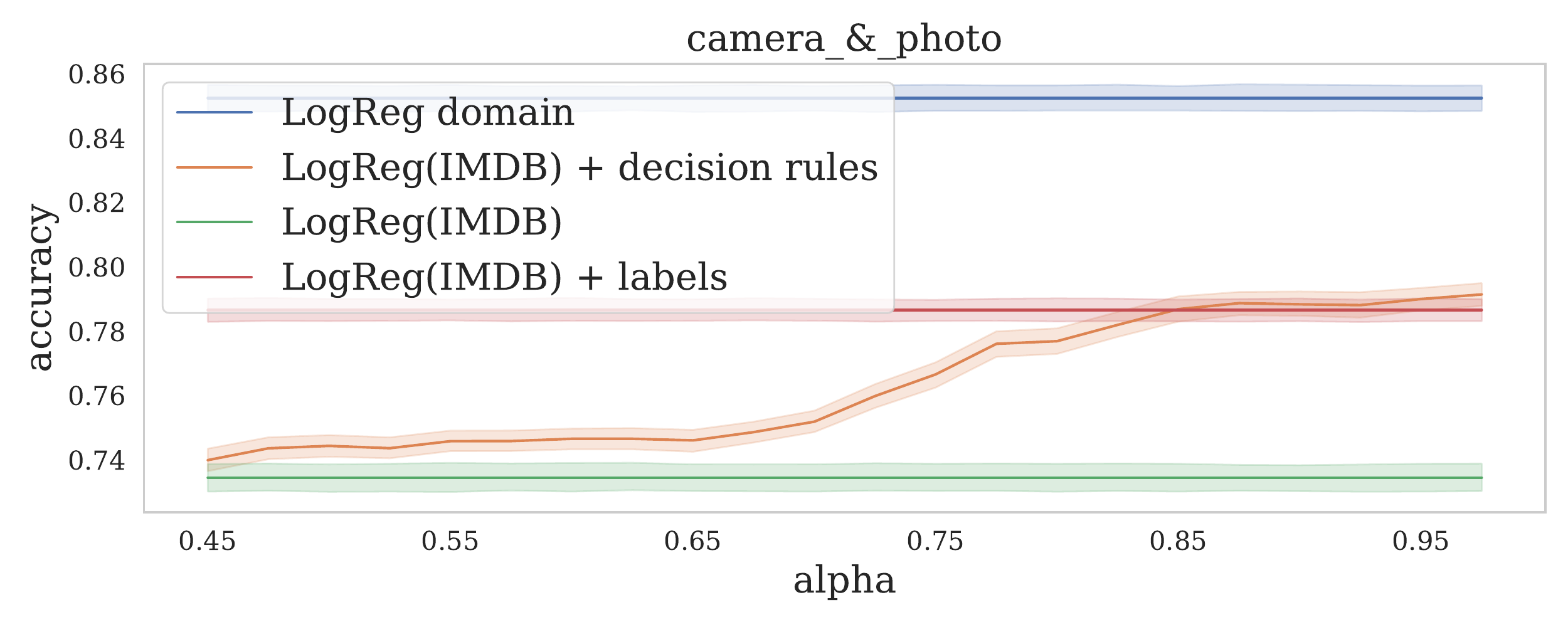}
}
\caption{The experimental results for the topics 1-6 (standard error is reported)}
\label{figure:sentiment_experiments_1}
\end{figure}

%====================================================================
\begin{figure}
\centering
\subfloat[]{
  \includegraphics[width=65mm]{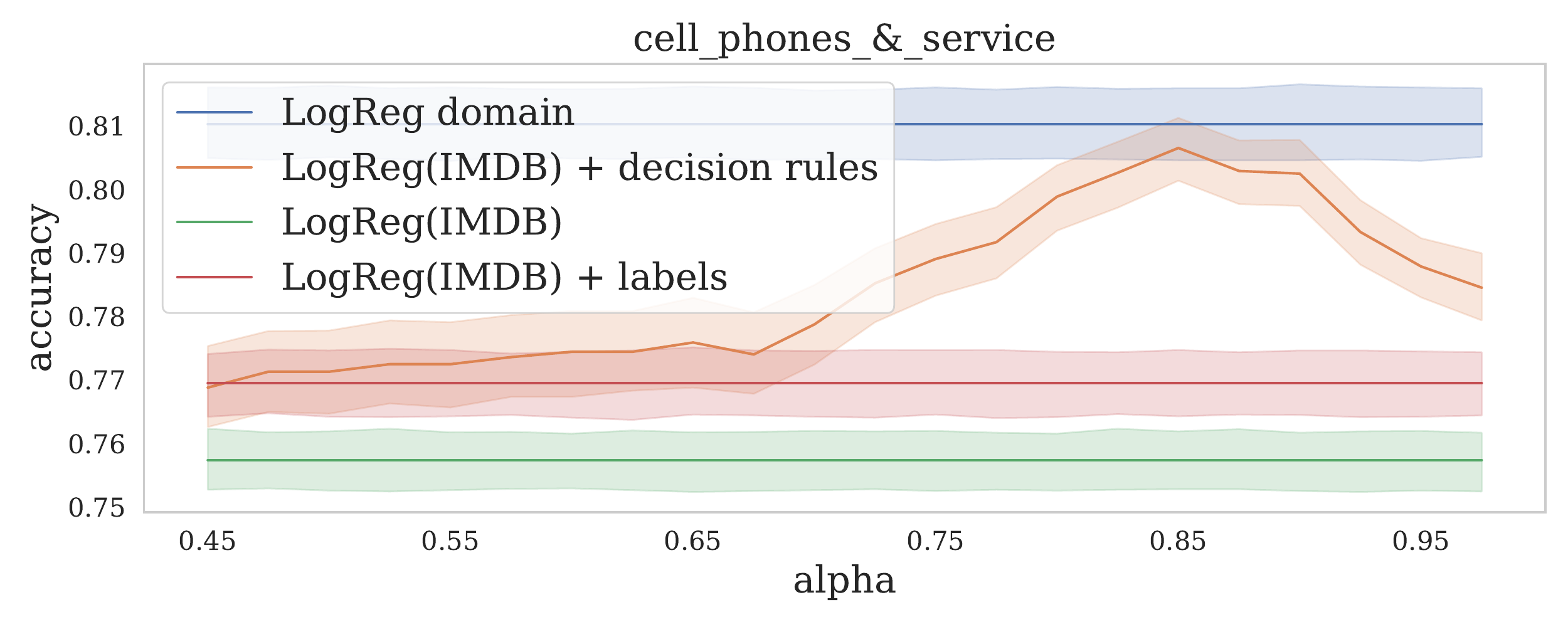}
}
\subfloat[]{
  \includegraphics[width=65mm]{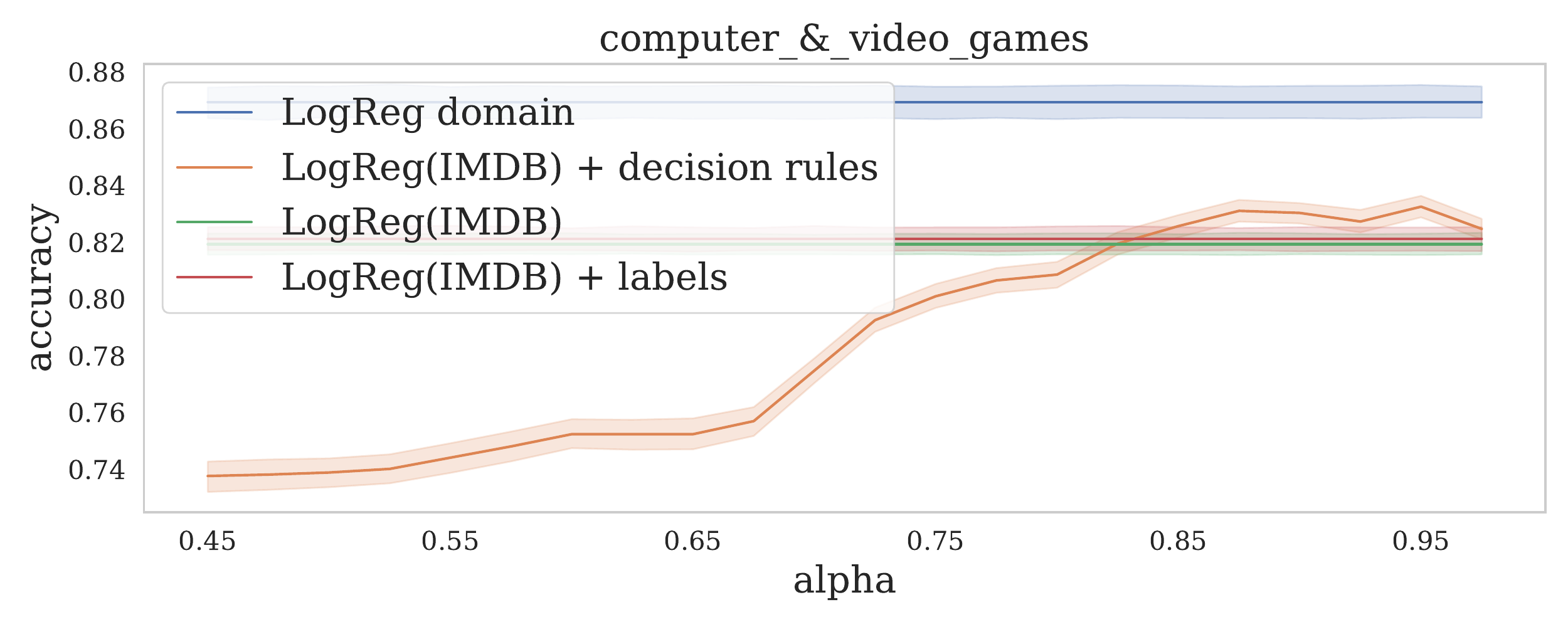}
}
\hspace{0mm}
\subfloat[]{
  \includegraphics[width=65mm]{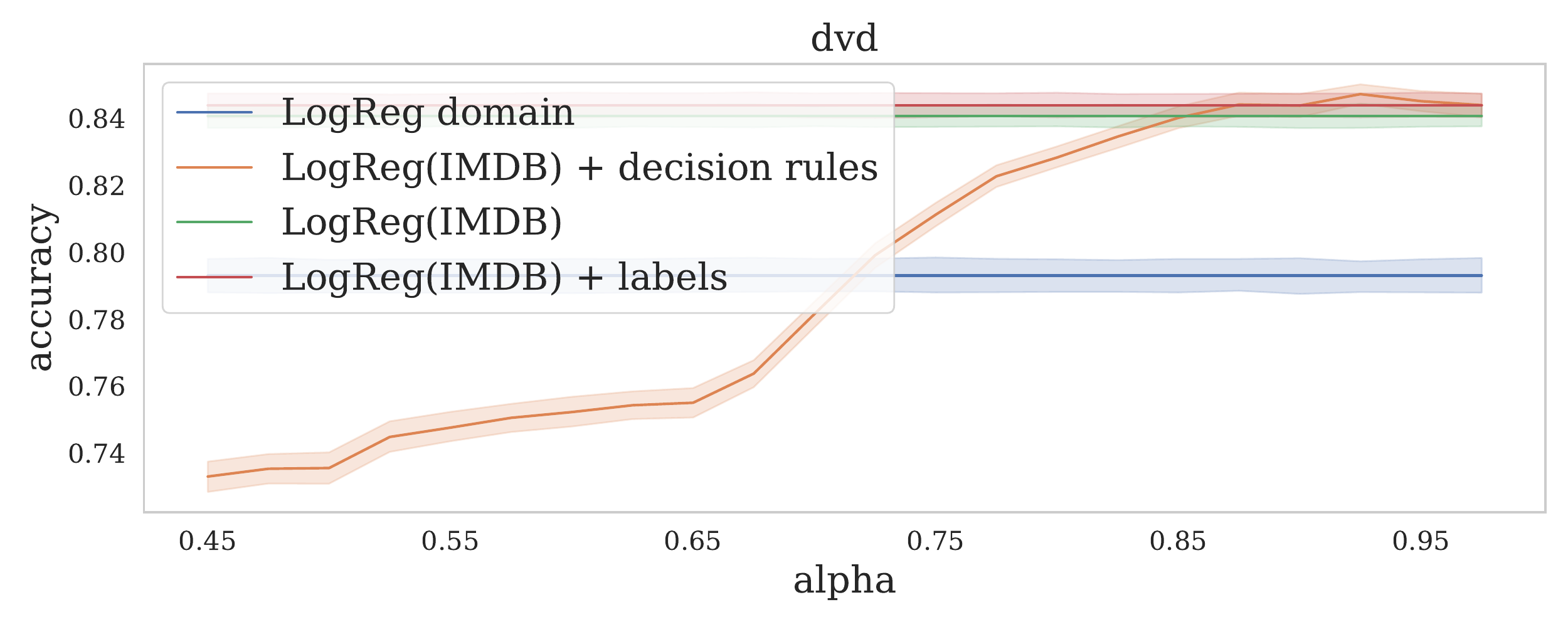}
}
\subfloat[]{
  \includegraphics[width=65mm]{images/sentiment_analysis/electronics.pdf}
}
\hspace{0mm}
\subfloat[]{
  \includegraphics[width=65mm]{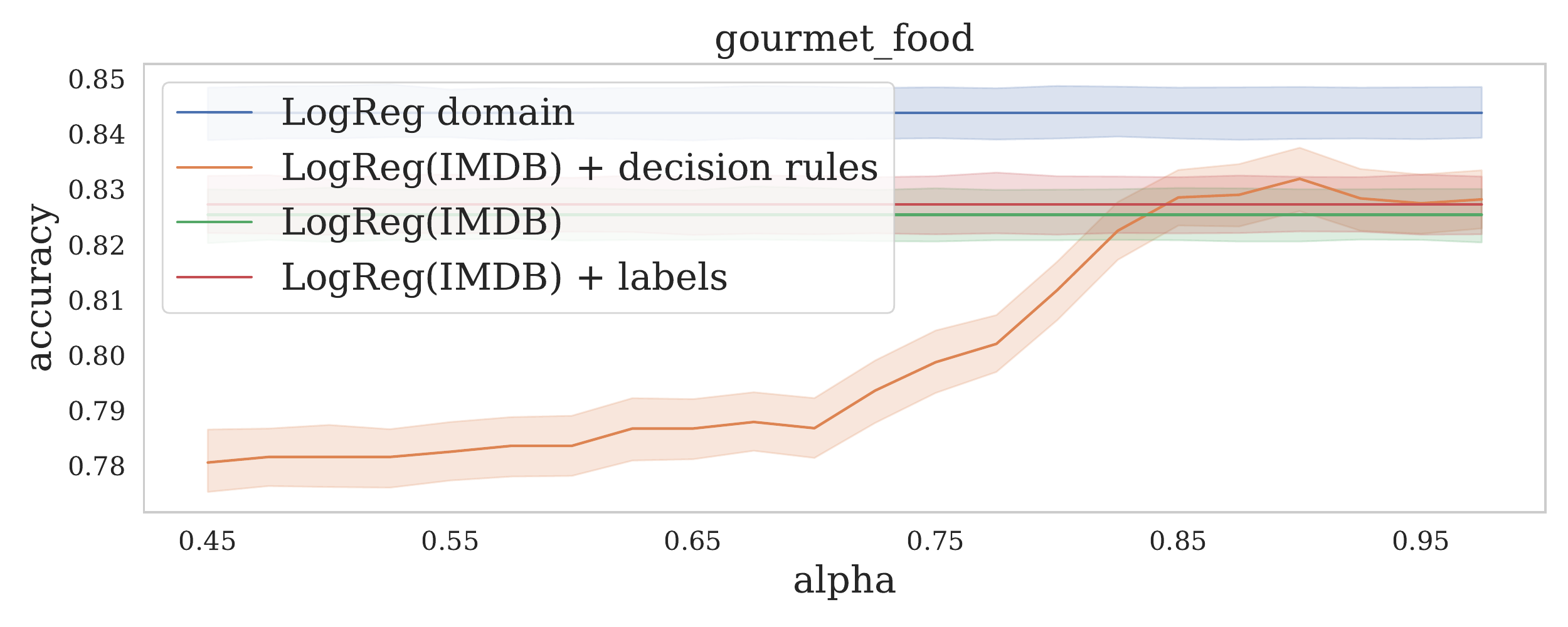}
}
\subfloat[]{
  \includegraphics[width=65mm]{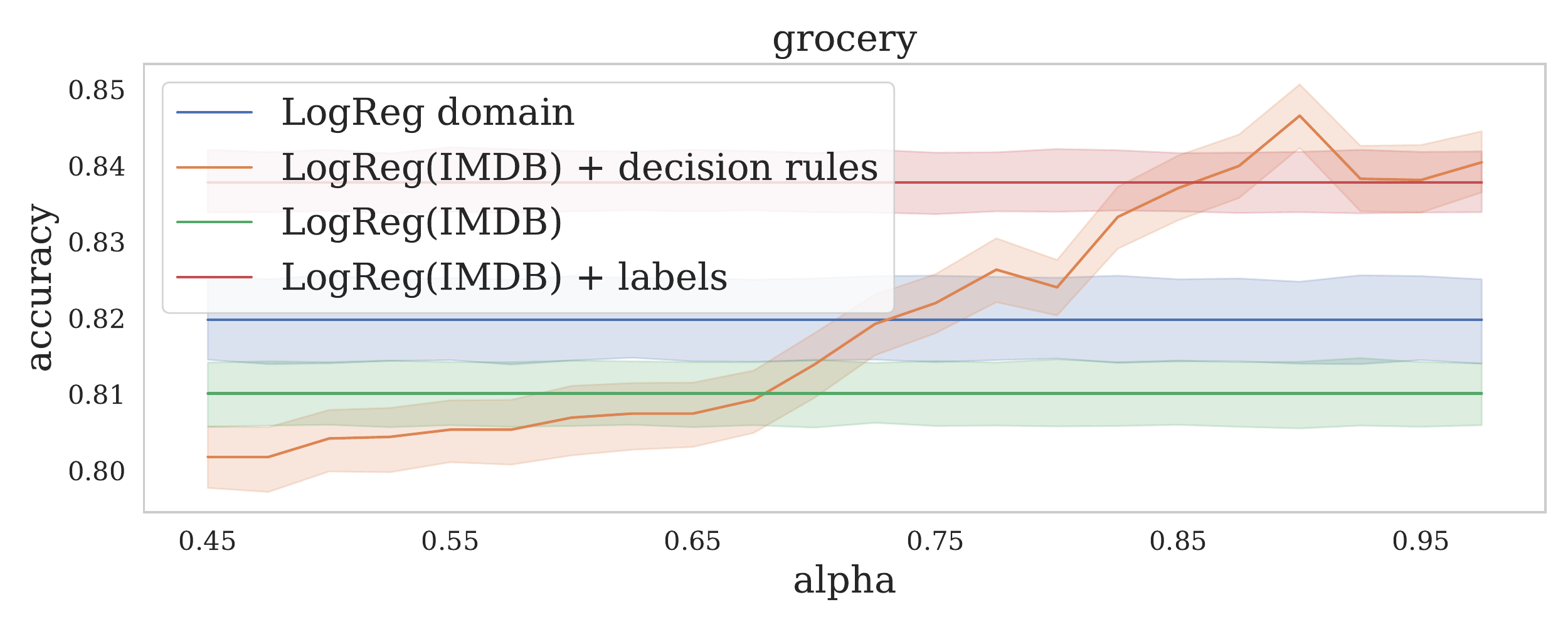}
}
\hspace{0mm}s
\subfloat[]{
  \includegraphics[width=65mm]{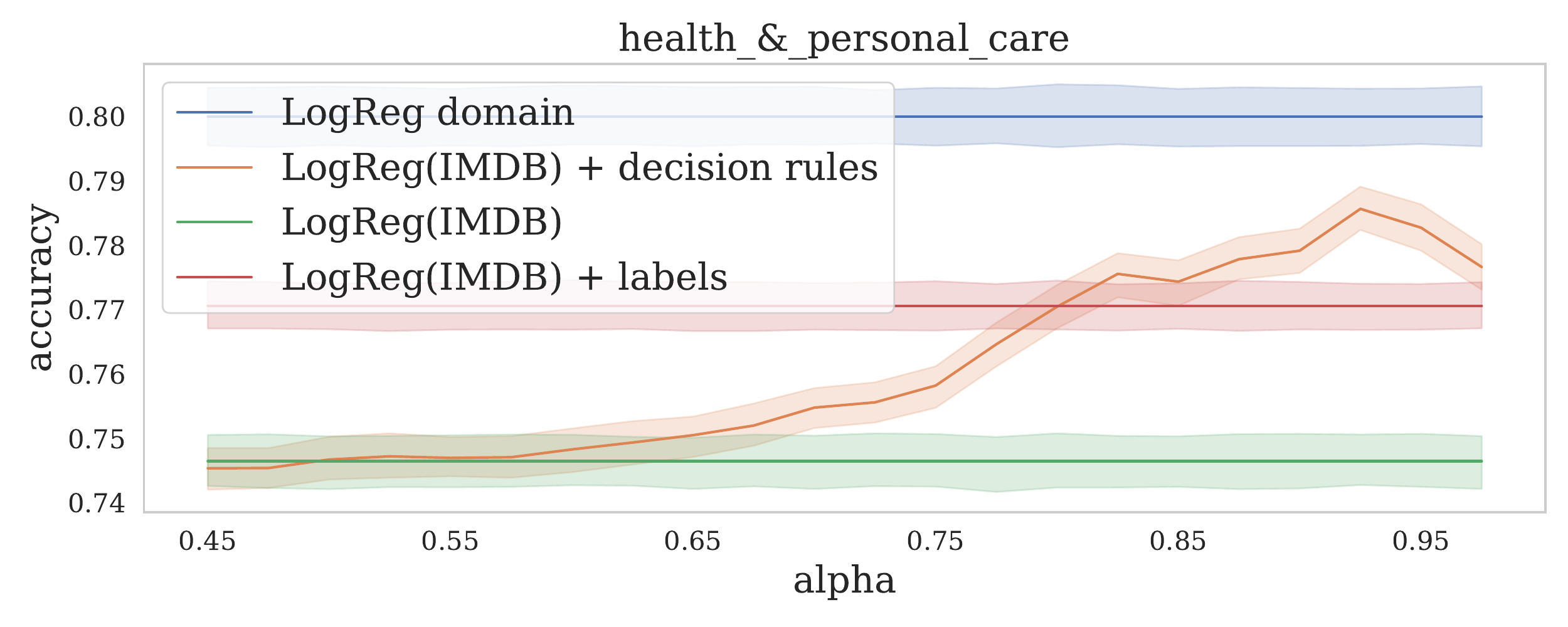}
}
\subfloat[]{
  \includegraphics[width=65mm]{images/sentiment_analysis/jewelry_and_watches.pdf}
}
\caption{The experimental results for the topics 12-22 (standard error is reported)}
\label{figure:sentiment_experiments_2}
\end{figure}

%===============================================================
\begin{figure}
\centering
\subfloat[]{
  \includegraphics[width=65mm]{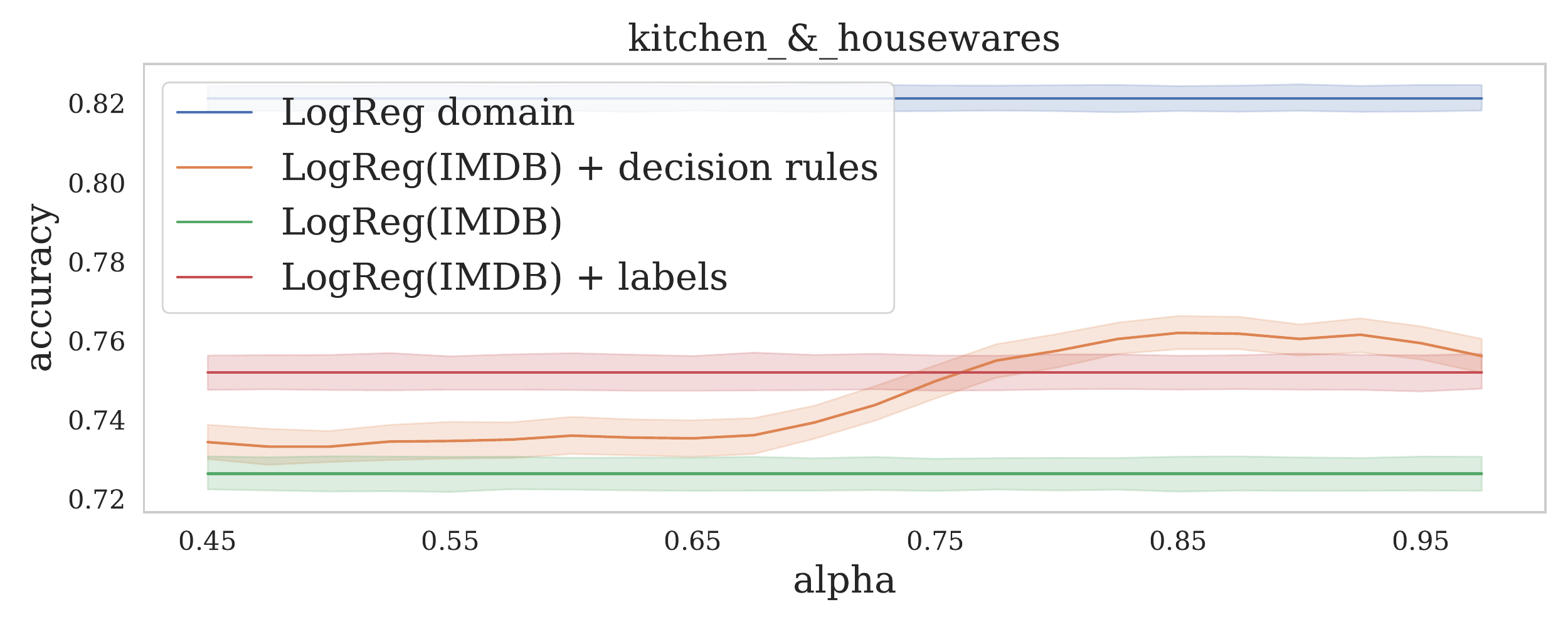}
}
\subfloat[]{
  \includegraphics[width=65mm]{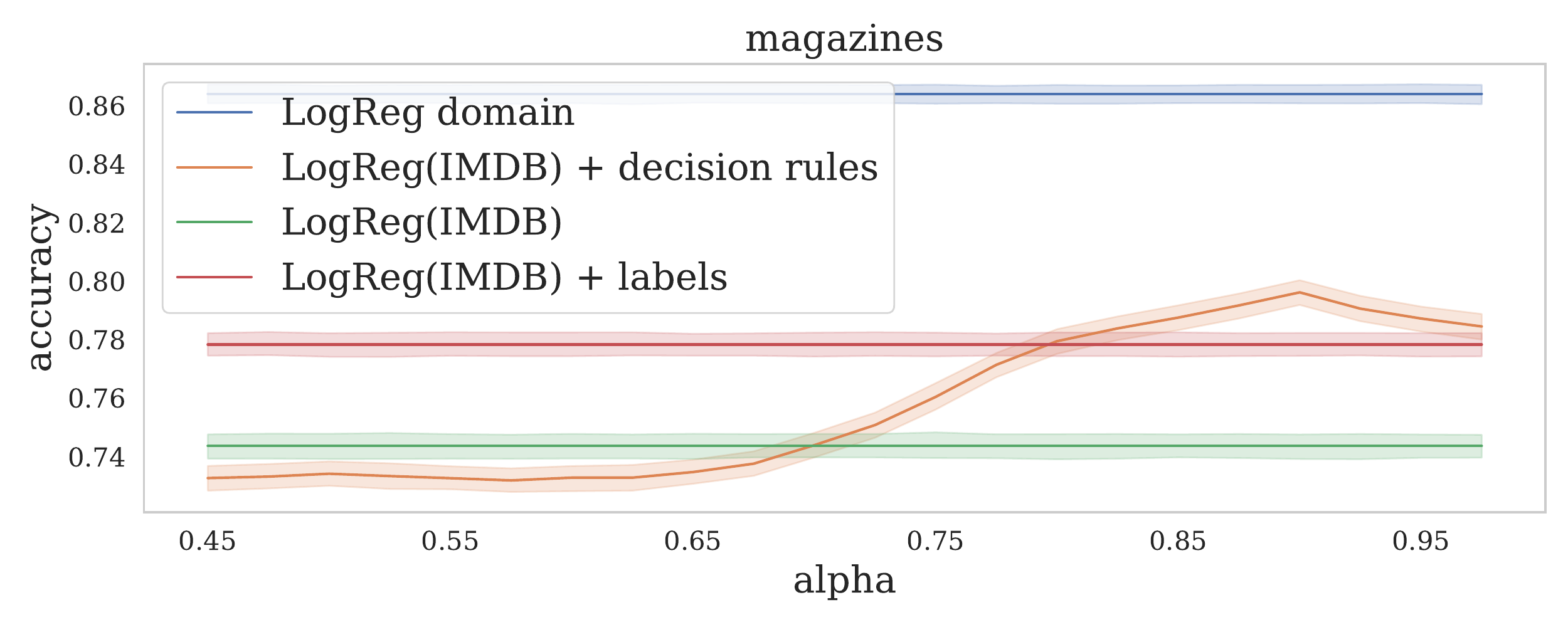}
}
\hspace{2mm}
\subfloat[]{
  \includegraphics[width=65mm]{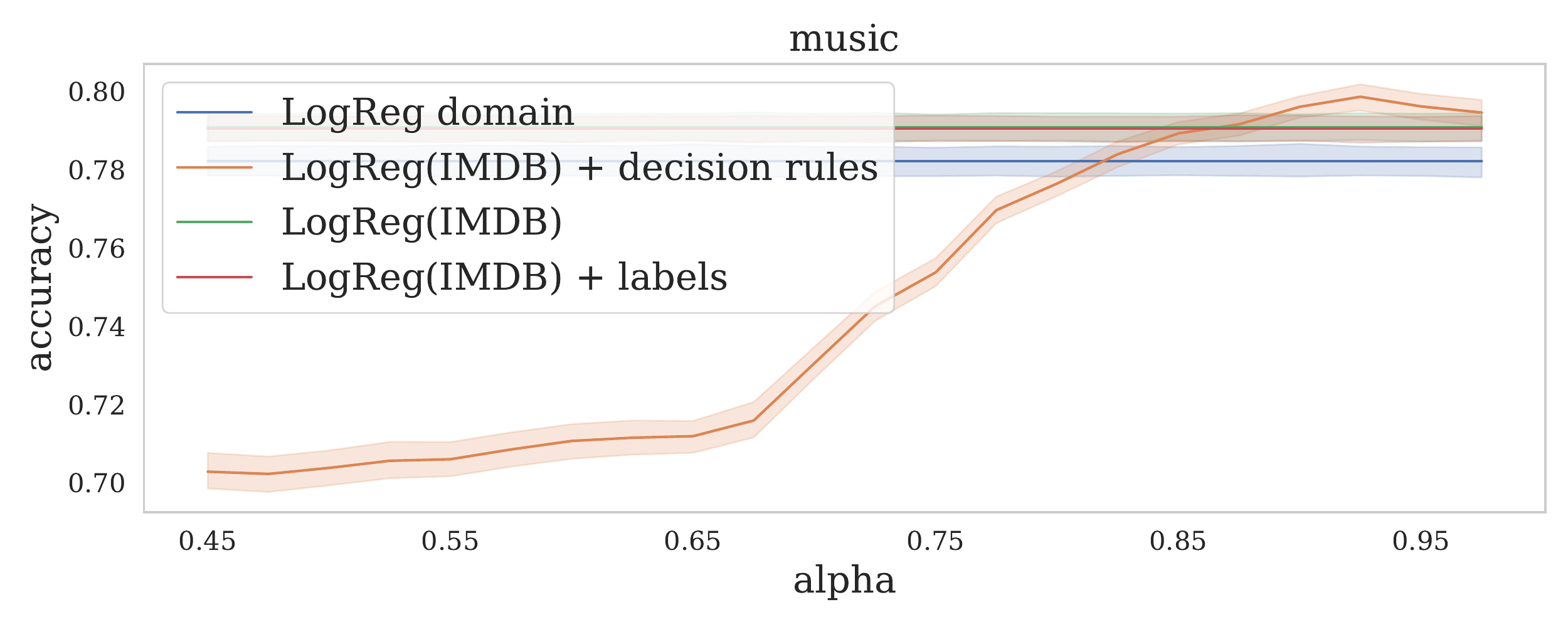}
}
\subfloat[]{
  \includegraphics[width=65mm]{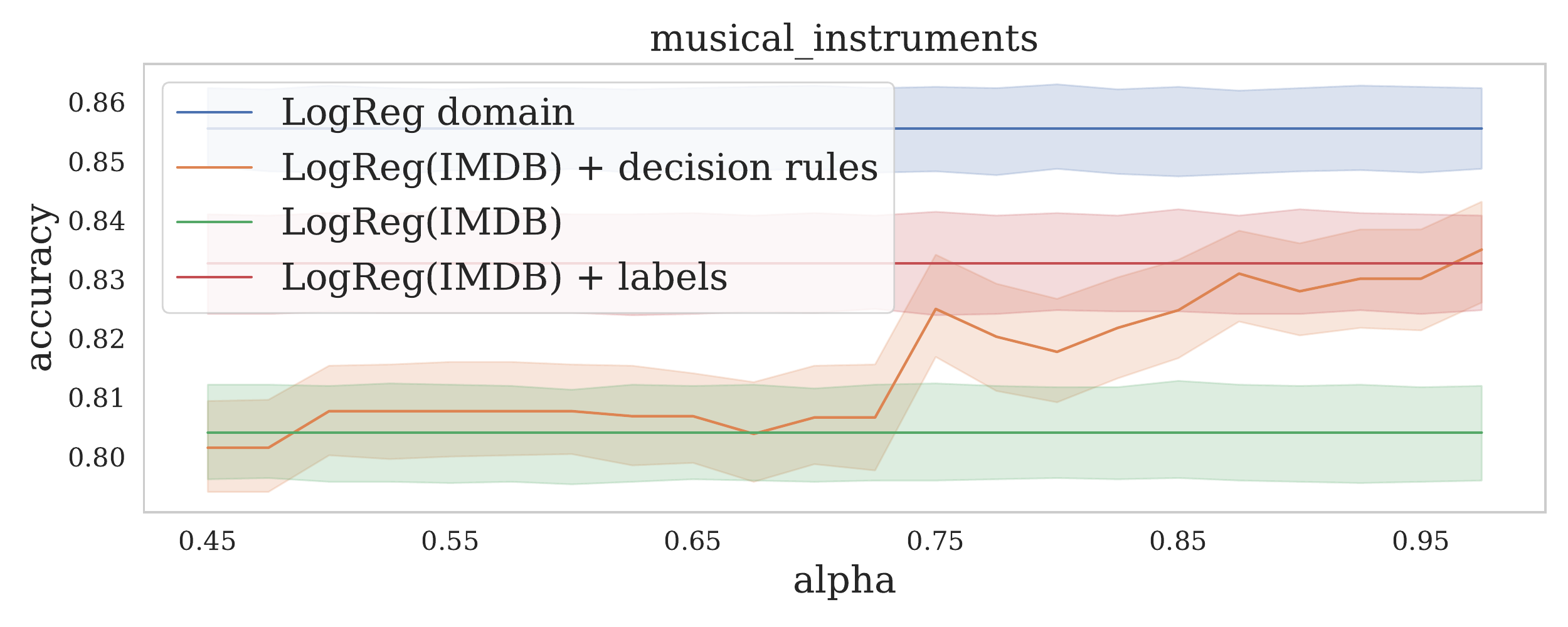}
}
\hspace{0mm}
\subfloat[]{
  \includegraphics[width=65mm]{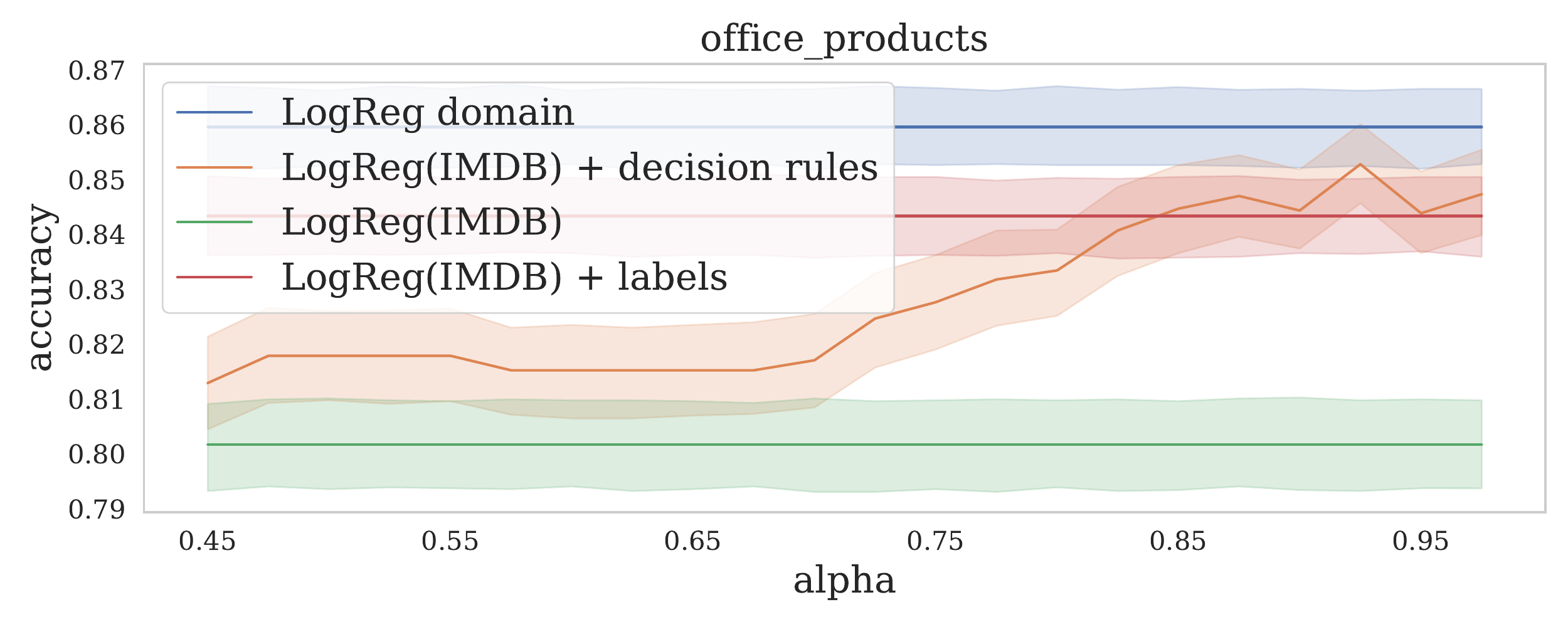}
}
\subfloat[]{
  \includegraphics[width=65mm]{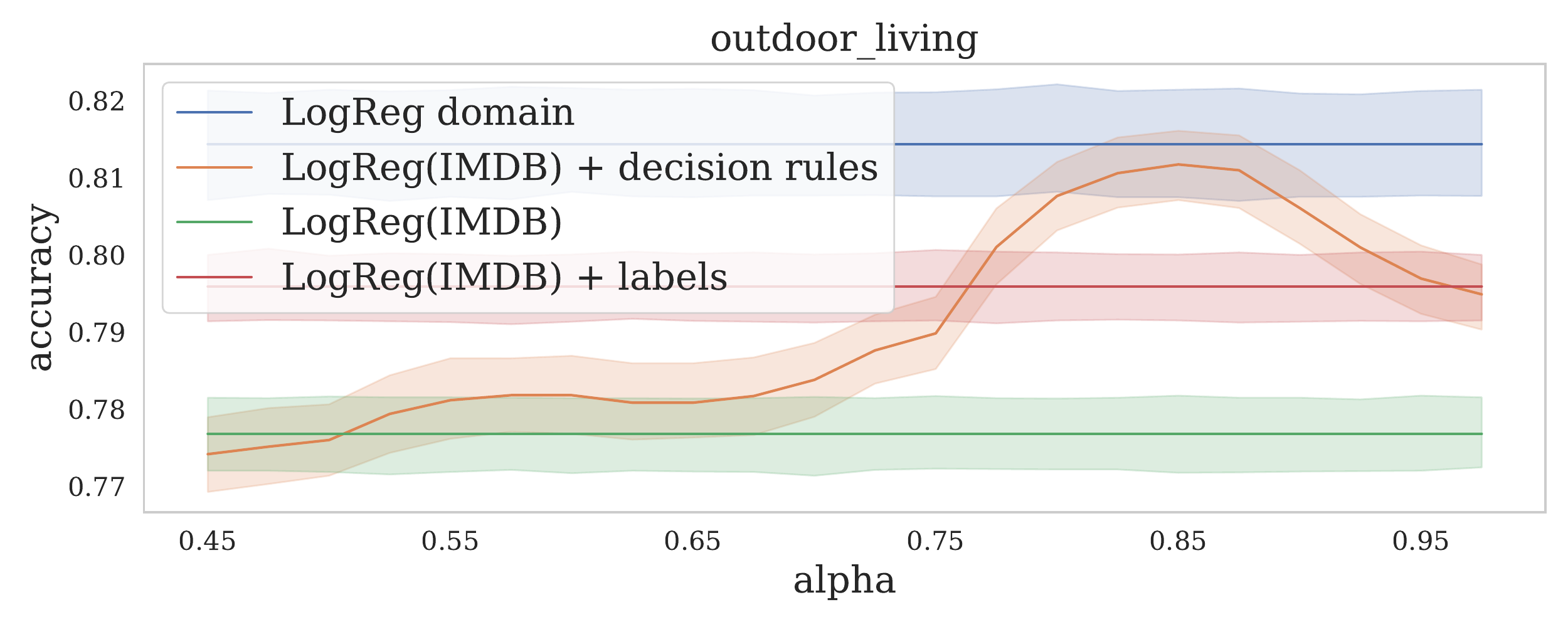}
}
\hspace{0mm}
\subfloat[]{
  \includegraphics[width=65mm]{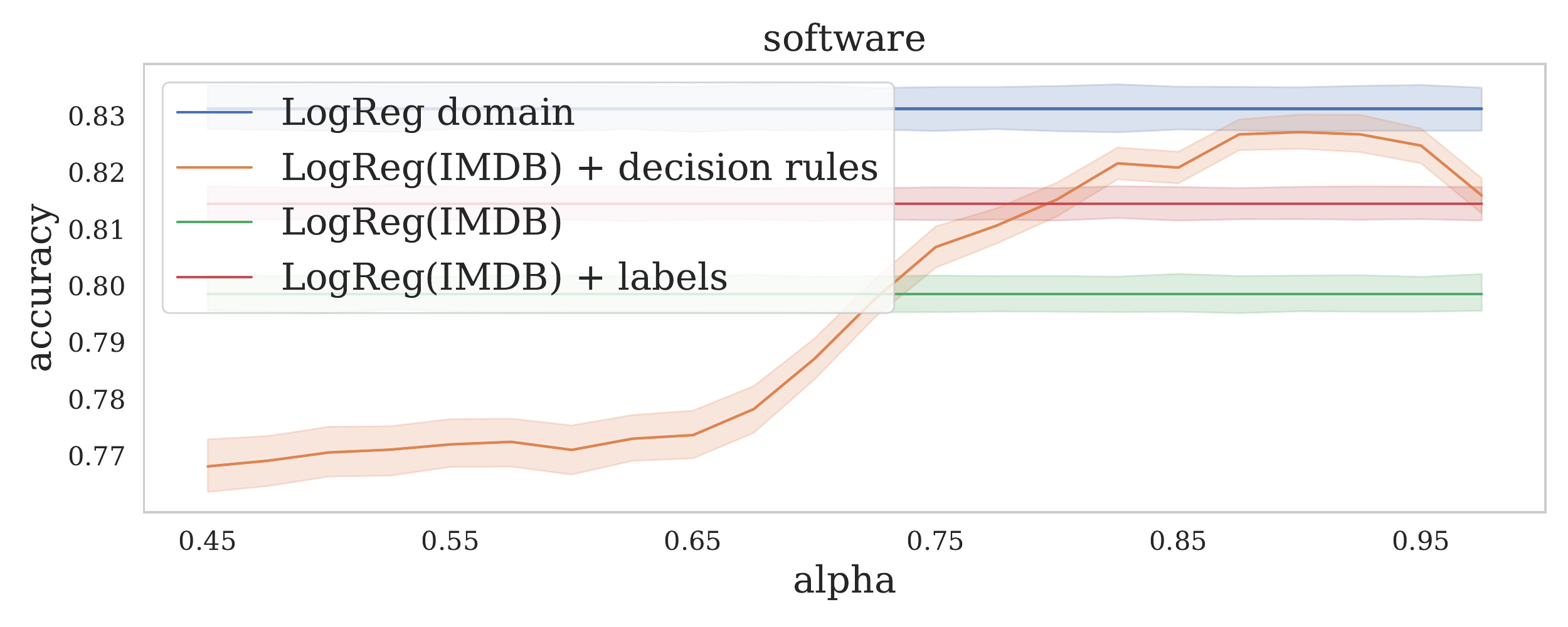}
}
\subfloat[]{
  \includegraphics[width=65mm]{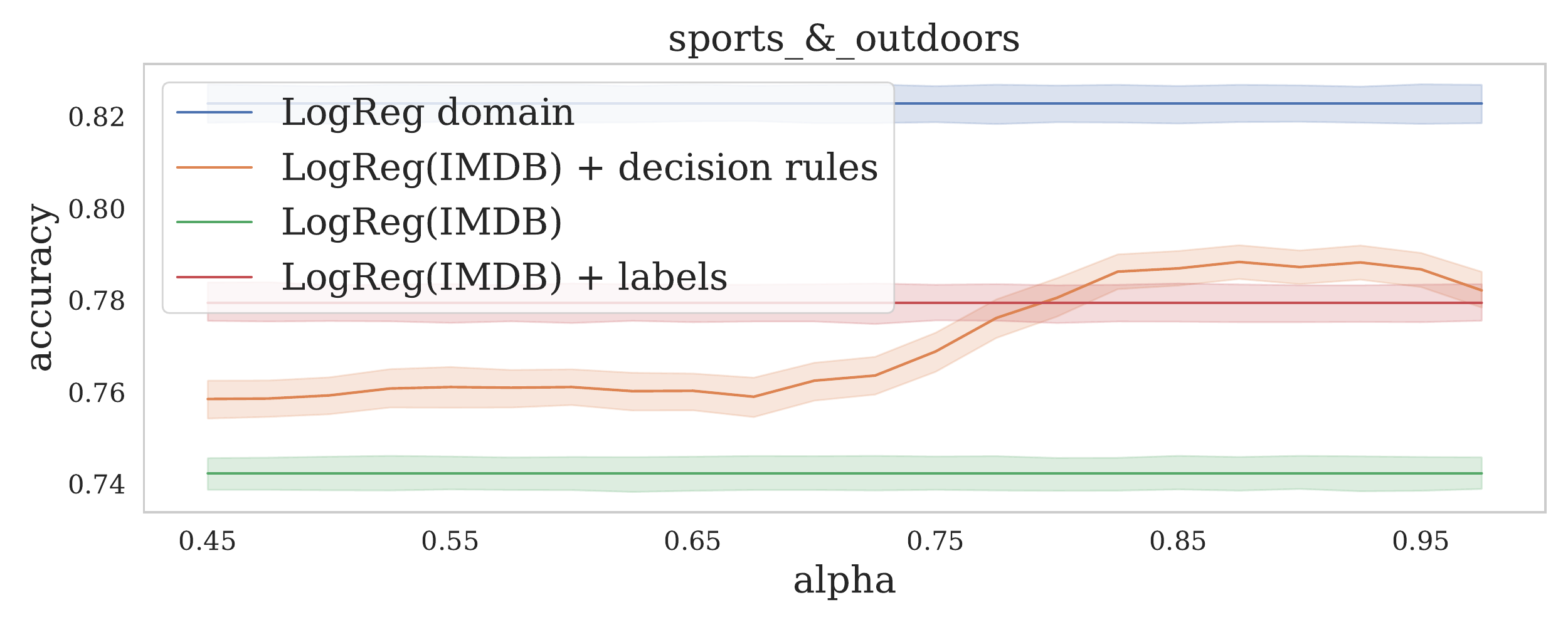}
}

\hspace{0mm}

\caption{The experimental results for the topics 22-25 (standard error is reported)}
\label{figure:sentiment_experiments_3}
\end{figure}

\begin{figure}
\centering
\hspace{0mm}
\subfloat[]{
  \includegraphics[width=65mm]{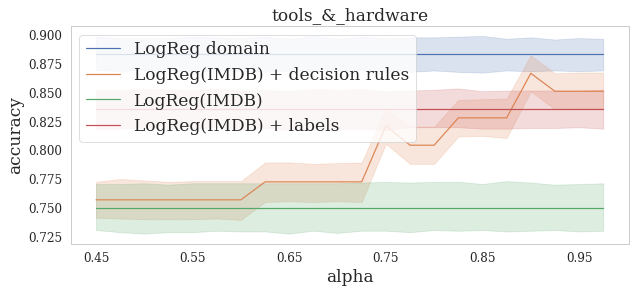}
}
\subfloat[]{
  \includegraphics[width=65mm]{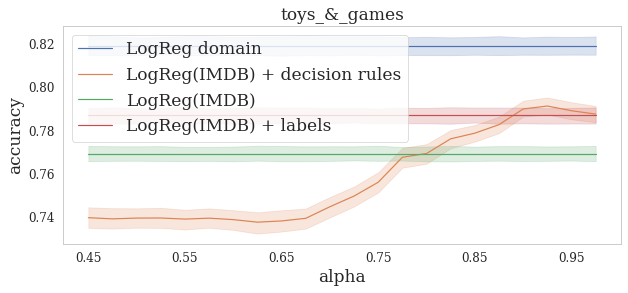}
}
\hspace{0mm}
\subfloat[]{
  \includegraphics[width=65mm]{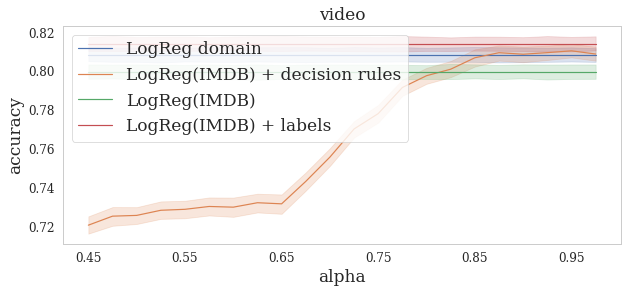}
}

\caption{The experimental results for the topics 22-25 (standard error is reported)}
\label{figure:sentiment_experiments_4}
\end{figure}

\fi

\end{document}